\documentclass[runningheads]{llncs}
\usepackage[T1]{fontenc}
\usepackage{graphicx}
\usepackage{booktabs}
\usepackage[misc]{ifsym}
\newcommand{\corr}{(\Letter)}
\usepackage{microtype}
\usepackage{amsmath}
\usepackage[caption=false]{subfig}
\usepackage{multirow}

\usepackage{tcolorbox}
\usepackage{listings}
\usepackage{siunitx}
\usepackage{glossaries}
\glsdisablehyper
\newacronym{AI}{AI}{artificial intelligence}
\newacronym{DNN}{DNN}{deep neural network}
\newacronym{MCU}{MCU}{microcontroller unit}
\newacronym{NAS}{NAS}{neural architecture search}
\newacronym{QAT}{QAT}{quantization-aware training}
\newacronym{PTQ}{PTQ}{post-training static quantization}
\newacronym{MOO}{MOO}{multi-objective optimization}
\newacronym{FLOP}{FLOP}{floating-point operation}
\newacronym{EH}{EH}{energy harvesting}
\newacronym{IC}{IC}{intermittent computing}
\newacronym{MAC}{MAC}{multiply-accumulate}
\newacronym{FRAM}{FRAM}{ferroelectric random-access memory}

\usepackage[hidelinks,colorlinks=false]{hyperref}
\usepackage{cite}
\usepackage{color}

\begin{document}

\title{Design-Time Optimization of Deep Neural Networks for Intermittent Learning on Microcontrollers}

\titlerunning{Design-Time Optimization of DNNs for Intermittent Learning on MCUs}

\author{Jakob Schubert\inst{1} \corr \and Maximilian Kasper\inst{1} \and Maximilian Linke\inst{1,2} \and Benedict Herzog\inst{3} \and  Mark Deutel\inst{1} \and Axel Plinge\inst{1} \and Dominik Seuss\inst{1,4} \and Christopher Mutschler\inst{1,5}}

\authorrunning{J. Schubert et al.}

\institute{Fraunhofer Institute for Integrated Circuits IIS, Nürnberg, Germany\\\email{\{jakob.schubert, maximilian.kasper, mark.deutel, axel.plinge\}@iis.fraunhofer.de} \and
Friedrich-Alexander-Universität Erlangen-Nürnberg (FAU), Erlangen, Germany\\\email{maxi.linke@fau.de} \and Ruhr University Bochum (RUB), Bochum, Germany\\\email{benedict.herzog@rub.de}
\and Technical University of Applied Sciences Würzburg-Schweinfurt (THWS), Würzburg, Germany \email{dominik.seuss@thws.de}
\and University of Technology Nuremberg (UTN), Nürnberg, Germany\\\email{christopher.mutschler@utn.de}}

\maketitle 
\setcounter{footnote}{0}

\begin{abstract}
We present a method for designing \glspl{DNN} for intermittent, energy-autonomous, on-device learning on \glspl{MCU}.
In mobile applications where the energy can run out, e.g., when solar-powered, executing \gls{AI} faces a technical issue as learning can be interrupted at any time.
Our approach combines a hardware-aware energy prediction model with \gls{MOO}, enabling offline \gls{DNN} optimization at the design stage without repeated deployment and online testing on the target \gls{MCU}.
Our proposed energy predictor estimates per-layer energy consumption for both \gls{DNN} inference and training, including the intermittent checkpointing overhead, based on implementation-specific compute and memory features extracted from the \gls{DNN} model.
We validate our approach using autoencoders for anomaly detection on a Cortex-M4 \gls{MCU}, where our predictor achieves a weighted absolute percentage error of 16.6\,\%, which is sufficient for reliable architecture selection under intermittency constraints.
As a result, this work bridges the gap between \gls{MOO}, automated \gls{DNN} design, deployment on energy-harvesting systems, and intermittent learning, truly enabling autonomous \gls{AI} at the edge.
\keywords{Intermittent Learning \and Energy Prediction \and Neural Architecture Search \and Microcontrollers.}
\end{abstract}
\glsresetall

\section{Introduction}

In many real-world scenarios, particularly in remote, inaccessible, or mobile deployment settings, continuous power supply is infeasible.
Instead, \gls{EH} enables autonomous operation by harvesting ambient energy (e.g., from light, vibration, or thermal gradients)~\cite{safaei2025eco}.
However, \gls{EH} introduces a fundamental challenge: energy availability is intermittent, non-deterministic, and often insufficient to complete an inference or training task in one continuous run~\cite{11261799}.

Deploying local \gls{AI} on resource-constrained edge devices requires careful design beyond accuracy considerations.
Critical constraints include limited Flash and RAM capacity, constrained computational resources, and the resulting impact on inference, latency, and energy consumption.
To explore this multidimensional design space, \gls{MOO}~\cite{deutel2025multi} and \gls{NAS}~\cite{qiao2024micronas} have become standard approaches.

\begin{figure}[t]
\includegraphics[width=\textwidth]{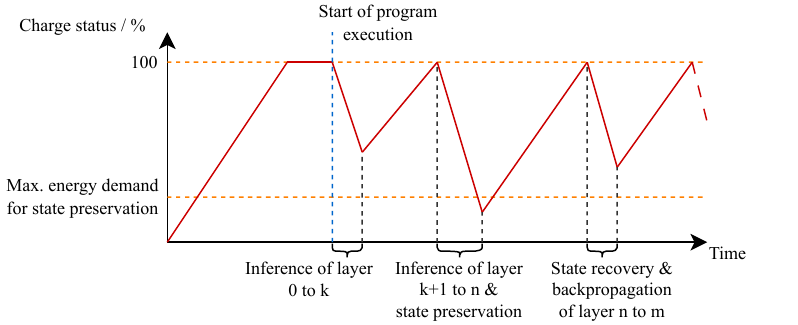}
\caption{Schematic example of energy availability during \gls{DNN} execution in an intermittent computing scenario. If the energy budget drops below a threshold, the state of the execution is preserved in non-volatile memory between layers. Once the energy budget has reset, the state is recovered and the execution continues.
} \label{fig:intermittent-learning}
\end{figure}

To deal with such intermittent energy conditions, the paradigm of \gls{IC} has emerged~\cite{lucia2017intermittent,umesh2021survey}.
In \gls{IC}, program execution is broken into checkpoints. Whenever the energy buffer reaches a sufficient level, computation resumes. If the buffer is about to deplete, the current program state is persisted to non-volatile memory at a predefined checkpoint, allowing safe interruption and later resumption.
In this work, we adapt \gls{IC} to \gls{DNN} inference and training by inserting checkpoints between layers in both the forward pass (inference) and backward pass (training), see Fig.~\ref{fig:intermittent-learning} for a schematic overview.
Consequently, the execution of a \gls{DNN} may span multiple energy cycles, making the energy budget no longer the sole constraining metric when designing \glspl{DNN}.
Instead, the energy consumption of the forward and backward operators of individual layers determine feasibility allowing the execution of bigger \glspl{DNN}.

This has profound implications for the design of \glspl{DNN}. A \gls{DNN} with lower total energy consumption may become infeasible if a single layer consumes more energy than the energy buffer can supply in one cycle, whereas a less energy efficient model, whose layer-wise energy profile fits the energy buffer, may instead end up as the viable option.

Consequently, we introduce a \gls{MOO} approach for \glspl{DNN} which considers intermittency constraints during search by using an energy prediction model that estimates layer-wise energy consumption of \glspl{DNN} without requiring physical deployment or measurements. 
Most \gls{MOO} approaches require the exploration of hundreds or thousands of \gls{DNN} architectures.
Thus, evaluating candidate models via real-world deployment and energy measurements is slow and error-prone. Instead, an accurate and fast to query energy prediction model is essential during optimization.

Furthermore, since \gls{DNN} weight updates (via uplink/downlink) are often unavailable or impractical in energy-harvesting systems, we expand our intermittent \gls{DNN} execution method to support on-device training of \glspl{DNN}. We do this by extending our energy prediction model to not only be able to predict energy consumption of forward passes of layers, but also their backward passes.

Summarizing our contributions, we propose a lightweight, portable energy prediction model trained on minimal empirical data, capable of estimating per-layer energy consumption for both inference and training. Furthermore, we demonstrate that this prediction model enables fast, hardware-aware \gls{MOO} at \gls{DNN} design time that explicitly considers feasibility under intermittent energy constraints, including buffer size limitations. Finally, we extend our prediction model to not only estimate the energy of forward passes, but also backward passes, enabling intermittent on-device training of \glspl{DNN} on energy-harvesting \glspl{MCU}.
\section{Related Work}
\label{sec:related_work}

Estimating the energy consumption of \gls{DNN} execution on resource-constrained \glspl{MCU} has been the subject of many studies.
A common approach is to base predictions on computational complexity, e.g., using \glspl{FLOP}, as a proxy for energy demand~\cite{lahmer2022energy,herzog2022resource}.
More precise prediction models incorporate memory access costs to improve accuracy, especially for memory-bound workloads~\cite{vsima2024energy1,vsima2024energy2,herzog2024greenpipe}.
These works emphasize data movement as the dominant energy factor, particularly in systems with multi-level memory hierarchies where memory bandwidth and access latency outweigh raw computation costs~\cite{puangpontip2022developing}.
Another line of work focuses on online energy profiling. Models are trained adaptively during operation using real-time measurements~\cite{10806703}.

Our approach draws inspiration from the feature-based modeling proposed by Puangpontip and Hewett~\cite{puangpontip2022developing}, but extends it in three key aspects.
\begin{enumerate}
\item We expand the feature space to cover a broader set of layer types beyond the commonly studied dense and convolutional layers to ensure high fidelity and hardware-awareness during energy prediction, critical for reliable optimization under strict energy, latency, and memory constraints.
\item We incorporate the backward pass into the energy model, enabling accurate energy prediction of on-device \gls{DNN} training, a capability absent in related work which focused solely on inference.
\item Our energy estimator is not derived from abstract mathematical descriptions of operators, but from measuring the actual implementation of the runtime and the hardware capabilities on the target \gls{MCU}.
\end{enumerate}

Recent work underscores the necessity of adaptability in edge \gls{AI} systems: models deployed in dynamic environments must be retrainable on-device to maintain performance over time~\cite{pittorino2026position}.
Several techniques have been proposed to make on-device training feasible under resource and energy constraints.
These include sparse weight updates to reduce memory and compute footprint~\cite{ren2021tinyol}, as well as specialized training algorithms for memory-limited \glspl{MCU}~\cite{lin2022device,deutel2024device}.
In this study we are using the training framework proposed by Deutel et al.~\cite{deutel2024device}, which enables full backpropagation on quantized \glspl{DNN} and which we integrate into the intermittent learning pipeline proposed in this paper.

\Gls{IC} poses unique challenges for learning systems.
Several recent efforts address these challenges: Nadalini et al.~\cite{nadalini2025multi} use online energy profiling to enable on-device training.
Fusco et al.~\cite{fusco2025device} reduce training energy via pruning and an early-exit strategy.
Other approaches leverage neuromorphic computing or specialized hardware to tolerate intermittent power~\cite{qazi2025intermittent}.
Our work directly contributes to this line of research by providing a design-time optimization tool for intermittent learning: instead of relying on runtime adaptation or hardware-software-co-design, we enable preemptive selection of architectures that minimize per-layer energy consumption, including intermittent checkpointing overhead, while meeting accuracy and memory constraints.
%
\section{Intermittent Learning Aware NAS}
\label{sec:concept}

\begin{figure}[t]
    \centering
    \subfloat[Energy prediction model.\label{fig:concept:dataset}]{%
        \includegraphics[width=.51\textwidth]{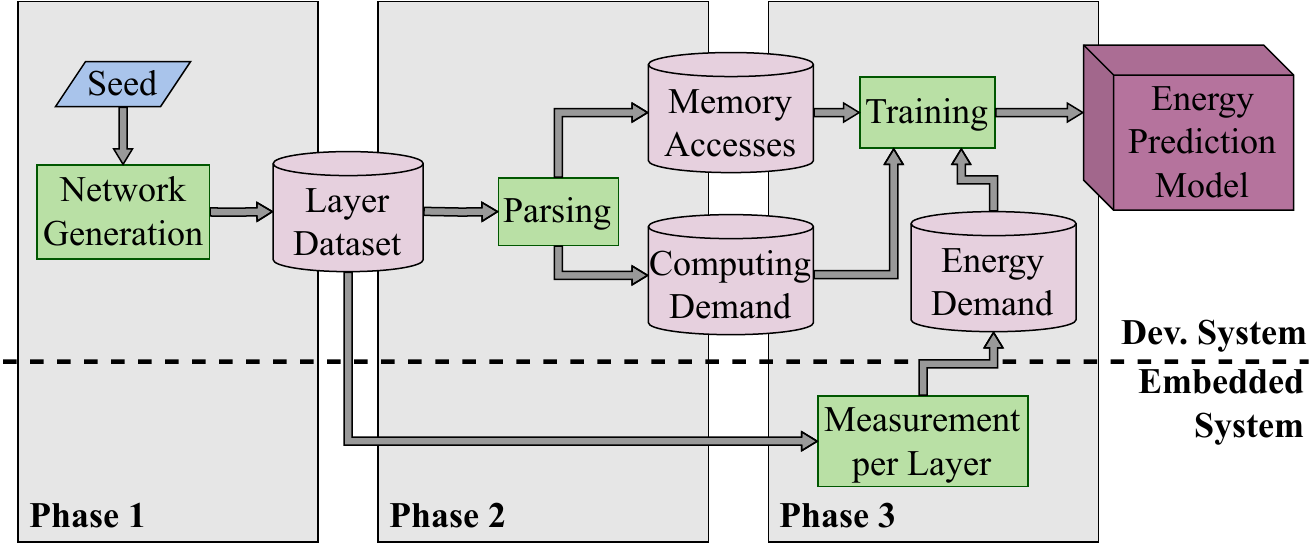}%
    }
    \hfill
    \subfloat[Multi-objective \gls{MOO} loop.\label{fig:concept:nas}]{%
        \includegraphics[width=.45\textwidth]{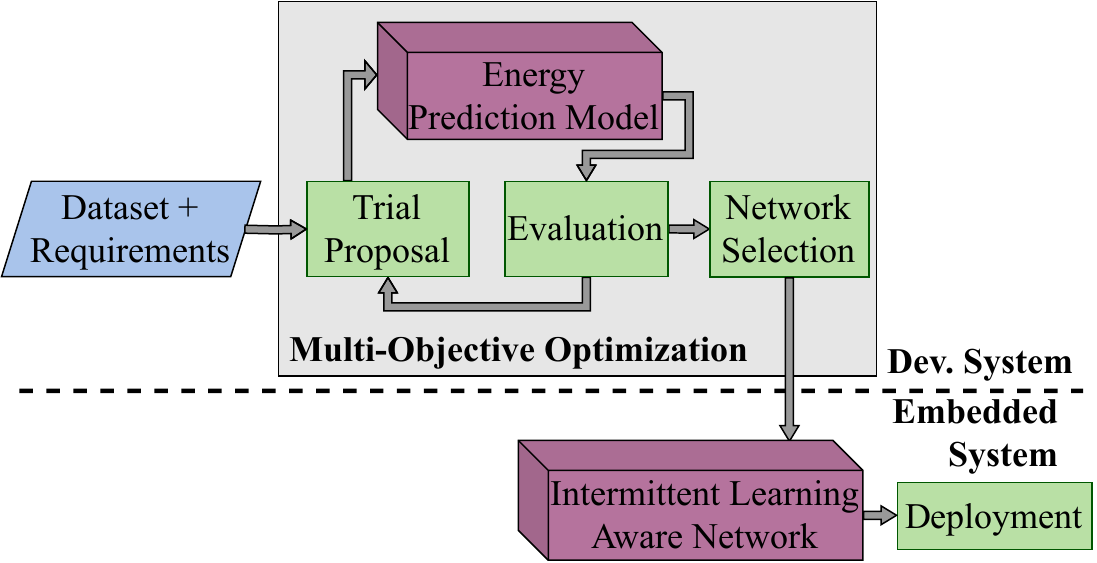}%
    }
    \caption{Schematic of the proposed method. Our pipeline is separated into construction of the energy-prediction model (Fig.~\ref{fig:concept:dataset}) and intermittency-aware \gls{MOO} (Fig.~\ref{fig:concept:nas}). The energy-prediction model constructed in the first phase can be used to quickly find and evaluate \gls{DNN} architectures and their estimated energy demand in the second \gls{MOO} phase.}
    \label{fig:concept}
\end{figure}

The selection of \glspl{DNN} for \gls{IC} requires a principled, \gls{MOO} framework that accounts not only for accuracy and resource constraints, but also for the unique constraints imposed by intermittent energy availability.
To this end, we propose a two-stage pipeline, see Fig.~\ref{fig:concept} for an overview.
\begin{enumerate}
\item The first stage handles automated acquisition of a dataset for fitting two lightweight, portable energy prediction models: one for the prediction of per-layer operator energy requirements and one for \gls{FRAM} read/write energy requirements (cf.~Sec.~\ref{sec:energypred}). 
\item The second stage performs an intermittency-aware \gls{MOO} using the two trained energy prediction models from the first stage (cf.~Sec.~\ref{sec:nas}).
\end{enumerate}

\subsection{Energy Prediction Models}
\label{sec:energypred}

We propose the usage of two linear regression models as energy predictors. Both models are trained with energy recordings from measurements on physical hardware.
The purpose of the first regression model is to estimate the energy consumption of a \gls{DNN}'s per-layer operators (both forward and backward passes), while the purpose of the second model is to estimate the energy consumption of saving and restoring a layer's state to and from \gls{FRAM}.
This allows for the quick verification of the feasibility of executing a \gls{DNN} under intermittent constraints, such as energy buffer size and maximum per-cycle energy budget, without physically deploying it. A dataset for training the two prediction models is constructed in three phases which we first briefly outline and then describe in more detail afterwards. A schematic of the phases can also be found in Fig.~\ref{fig:concept:dataset}.

\begin{enumerate}
\item \emph{Phase 1: Layer Sampling and Architecture Generation.}
We sample from sets of primitive layer types (e.g., convolutions, fully connected, max. pooling, etc.) and corresponding parameter ranges to build a composition of layers with broad coverage of search space.
\item \emph{Phase 2: Feature Extraction.}
We extract compute (i.e., \gls{MAC} equivalent operations) and memory (i.e., RAM accesses) as independent features for per layer energy regression as well as memory (i.e. FRAM accesses) for energy prediction of \gls{DNN} state preservation.
\item \emph{Phase 3: Empirical Measurement and Regression Model Training.} We compile and deploy every sampled \gls{DNN} on the target \gls{MCU}. Using power profiling, we measure the energy consumption of each layer for inference (forward pass) and training (backward pass) as well as for its persistent storage to \gls{FRAM} in case of intermittency. Afterwards the linear regression models are trained using the previously defined features and the corresponding measured energy values as dependent variable for each sampled \gls{DNN} configuration.
\end{enumerate}

\begin{figure}[t] 
    \centering 
    \includegraphics[width=\textwidth]{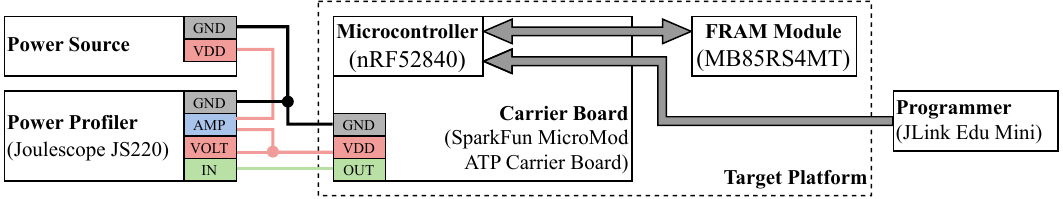}
    \caption{Hardware setup for energy measurements. The system was used to automatically deploy \glspl{DNN} on the target platform, measuring the power consumption and inference time and deriving the energy consumption from it. To quantify the energy consumption for model state persistence the write/read to/from a \gls{FRAM} via SPI was measured.} 
    \label{fig:hardware_setup}
\end{figure}

\paragraph{Layer Sampling and Architecture Generation.}
Using a fixed random seed, we first generate a diverse set of \gls{DNN} architectures by composing configurable primitive layers.
We then export all sampled \glspl{DNN} to ONNX format for portability. 
For execution on the target \gls{MCU} we use a proprietary runtime which includes a tool for preemptive ONNX-to-C code conversion, static memory allocation for weights, activations, and gradients at compile time, and hardware optimized operator implementations (e.g., using DSP instructions).

\paragraph{Feature Extraction.}
We implement a layer-level feature extractor to estimate the compute and memory features from a given ONNX file. For each forward and backward operator of a supported layer, the extractor queries the relevant tensor dimensions and attributes from the ONNX data structure. 
In addition, it takes hardware-specific configuration options into account, such as whether DSP instructions or hardware floating-point support are available on the targeted \gls{MCU}. 
As these hardware-specific options influence which implementation of an operator is used on the target \gls{MCU} they also change the resulting compute and memory estimates.

Consequently, the estimator does not analyze the code running on the \gls{MCU} one instruction at a time to estimate overall compute and memory requirements, but rather abstracts each layer operator into features, such as the depth and dimensions of an operator's loop nest or the accessed data structures.

Exemplary, for a fully connected layer with batch size $B$, input dimension $I$, and output dimension $O$, the total number of \gls{MAC} equivalent operations is estimated as stated in Eq.~\eqref{eq:maceqop}.
\begin{equation}
    N_{\mathrm{MAC}} = B \cdot I \cdot O
    \label{eq:maceqop}
\end{equation}
Furthermore, memory-access is estimated assuming that the input activation tensor contains $B \cdot I$ elements, the weight matrix contains $O \cdot I$ elements, the bias vector contains $O$ elements, and the output tensor contains $B \cdot O$ elements.
Multiplying each tensor access count by its corresponding bit width results in Eq.~\ref{eq:memaccess}, where $b_{\mathrm{in}}$, $b_{\mathrm{w}}$, $b_{\mathrm{bias}}$, and $b_{\mathrm{out}}$ denote the bit widths of the input activations, weights, bias values, and output activations, respectively.
\begin{equation}
\begin{aligned}
    N_{\mathrm{IObits}}
    =
    &\; B I \cdot b_{\mathrm{in}}
    + O I \cdot b_{\mathrm{w}} \\
    &+ O \cdot b_{\mathrm{bias}}
    + B O \cdot b_{\mathrm{out}},
\end{aligned}
\label{eq:memaccess}
\end{equation}

\gls{IC} checkpoints are positioned between each layer in both the forward and backward pass. This ensures the layer's state is preserved or recovered in the event of intermittency.
As feature for energy demand prediction in state preservation in \gls{FRAM}, the number of bytes to write/read is taken into account.

We apply the same principle as shown above for fully connected layers to estimate compute and memory of the forward and backward passes of all supported layer types.

\paragraph{Empirical Measurement and Regression Model Training.}
The energy regression feature extraction is designed to be applicable to any microcontroller platforms, only the energy measurments need to be repeated once for new hardware platforms.
We performed all energy measurements using an automated setup, see Fig.~\ref{fig:hardware_setup}, where each sampled \gls{DNN} is first compiled, then flashed via a JLINK debug probe, and finally executed layer-by-layer on the target \gls{MCU}. In this work, we used an nRF52840 Cortex-M4 \gls{MCU} with \SI{64}{MHz} clock speed, \SI{256}{kB} of SRAM, \SI{1}{Mb} Flash, an FPU, and ARM's DSP extension. 
To isolate energy consumption per layer, the \gls{MCU} was programatically put into its sleep mode between each layer and GPIO pins were toggled at each layer's entry/exit to mark their execution. 
We used a Joulescope JS220 power profiler to measure supply voltage and current of the system including the attached \gls{FRAM} module. The Joulescope can also automatically synchronize its measurements with recorded GPIO events which allows for precise temporal alignment between recorded power and layer execution windows. 
For persistent memory operations, a 4-Mbit SPI-connected \gls{FRAM} (MB85RS4MT) was used, and energy for reads/writes of 10 to \SI{300}{kB} was recorded analogously.

We use the extracted features and energy measurements to train the two regression models to predict layer-wise energy consumption.
Compute and memory features are designed to universally enable energy prediction of layers in inference and training modes simultaneously even if energy characteristics differ substantially (e.g., gradient computation and parameter updates introduce additional memory writes and compute overhead).
While the resulting predictors are specific to a certain runtime and target \gls{MCU}, they are compact and portable. They also require no deployment on physical hardware, making them suitable for integration into any existing NAS framework.

\subsection{Intermittency-Aware Multi-Objective Optimization}
\label{sec:nas}

We show an overview of our \gls{MOO} algorithm using the energy prediction models from Sec.~\ref{sec:energypred} in Fig.~\ref{fig:concept:nas}. The algorithm requires a dataset for \gls{DNN} training and a set of constraints and optimization objectives as input.
Objectives typically are to maximize validation accuracy while minimizing RAM/ROM footprint, total energy, and, crucially for intermittency, restricting the peak energy demand per layer. Constraints are derived from the limits of the targeted \gls{MCU}, e.g., to avoid overflowing its SRAM or Flash.

During optimization, new \gls{DNN} candidates are proposed iteratively, and their energy objective values are evaluated using the energy prediction models.
Specifically, each candidate's layer-wise computation and memory features are extracted, handed to the predictors, and used to infer the corresponding total energy consumption.
If a stopping criterion is met, the set of candidate architectures is filtered to retain only feasible architectures (e.g., minimum acceptable accuracy, maximum memory size, or maximum per-layer energy budget).
From this set, the Pareto-optimal solutions across accuracy, memory, and energy are selected.

By using this method, the entire \gls{MOO} can run entirely offline. No code is deployed or measured on the target \gls{MCU} during optimization. Once found, the selected final \gls{DNN} is deployed with the option of being safely updated later on via on-device training (as the backward pass has been factored in by the energy prediction models), enabling continuous adaptation in an intermittent setting.

As a result, our approach enables exploration of thousands of architectures optimizing for intermittency constraints omitting the necessity of deployment and measurements on the target platform, drastically accelerating design space exploration while preserving physical fidelity.
\section{Evaluation}
\label{sec:results}

\begin{figure}[t]
    \centering
    \subfloat[Network layer regression.\label{fig:regression_layers}]{%
        \includegraphics[width=.48\textwidth]{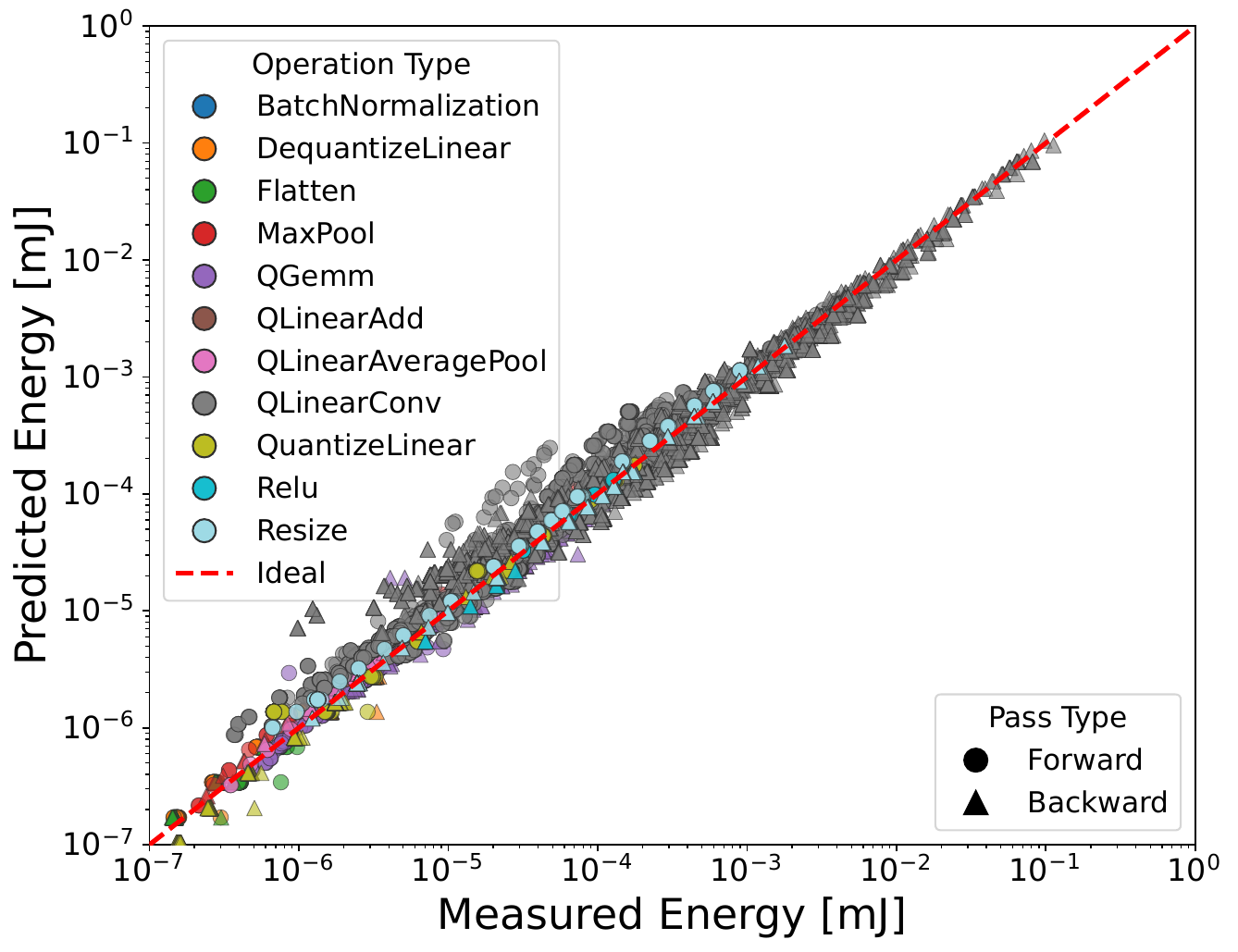}%
    }
    \hfill
    \subfloat[\gls{FRAM} regression.\label{fig:regression_fram}]{%
        \includegraphics[width=.48\textwidth]{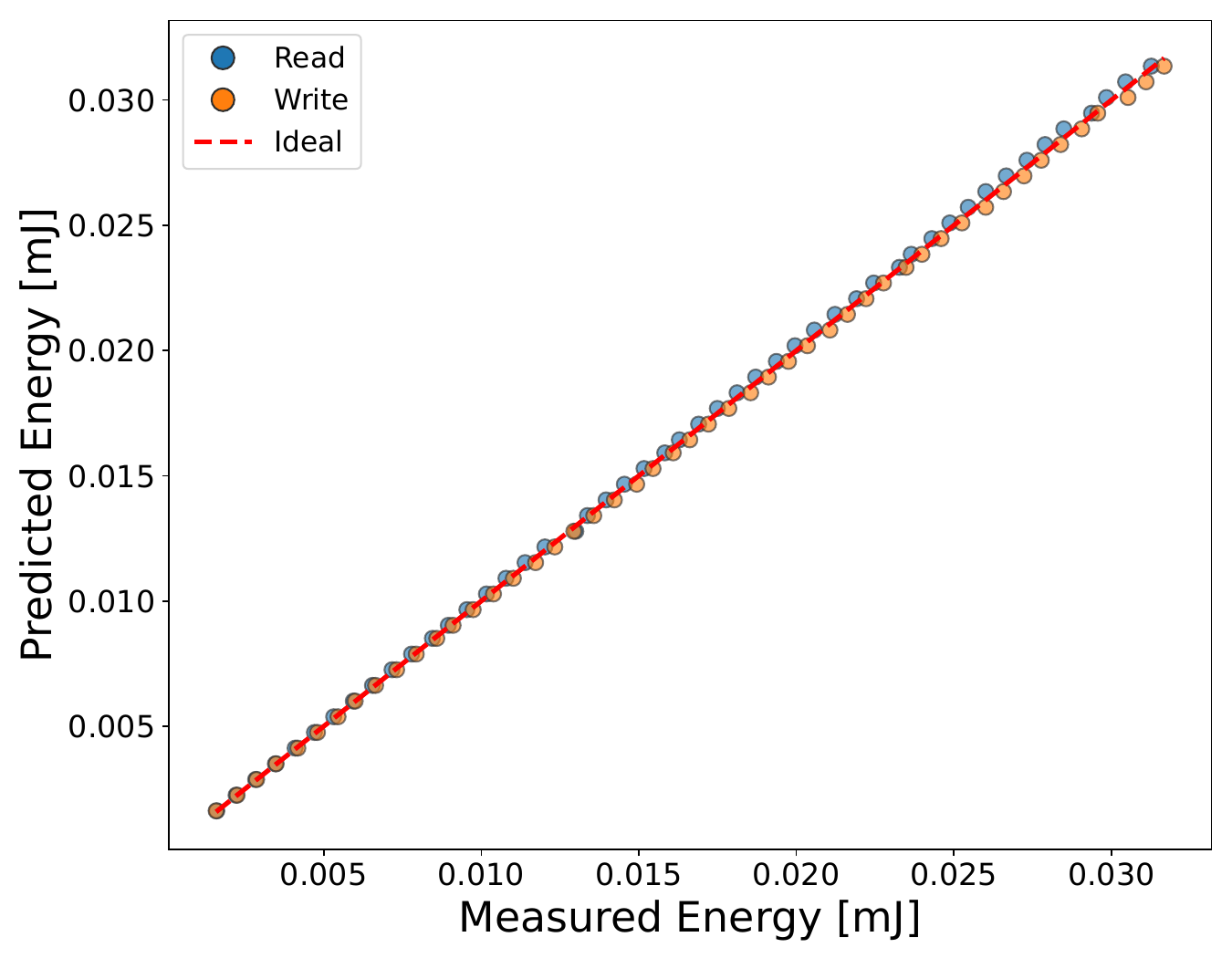}%
    }
    \caption{Energy estimator regression results for energy prediction.}
    \label{fig:regression_combined}
\end{figure}

\subsection{Performance of the Energy Prediction Models}

We first evaluate the two energy prediction models we proposed in Sec~\ref{sec:concept}. Initial experiments showed a strong linear correlation between the modeled compute/memory features and the measured energy across all layer types.
This motivated our design decision to use a single linear regression model for all layers, rather than per-layer-type models.
The key advantage of this unified approach is extensibility: new layer types can be added to the search space without requiring additional physical measurements, provided their compute and memory characteristics are analytically estimated.
In addition, the simplicity and explainability of linear regression facilitate extending the set of supported layers without retraining, thereby preventing overfitting to the training set.

Our training dataset consists of around 5000 individual layer instances across 11 layer types (e.g., dense, convolutional, pooling, activation), each measured for both inference and training mode.
Using this dataset, we fit the two linear regression models that map the aggregated feature vector (\gls{MAC} equivalent operations and I/O bits) to the measured energy and show the results in Fig.~\ref{fig:regression_combined}.
The predictions of both the layer regression model in Fig.~\ref{fig:regression_layers} and the \gls{FRAM} regression model in Fig.~\ref{fig:regression_fram} show a strong linear correlation across all layer types, execution modes, and read/write cycles.

\begin{table}[t]
\centering
\caption{Average error values (MAE, MAPE, WAPE) of the layer energy estimator for forward (fwd) and backwards (bwd) passes of sampling groups of up to a hundred measurements per layer (\#). Also shown are the average FRAM energy estimator error values for reading and writing operations to non-volatile FRAM.}
\label{tab:result_metrics}
\begin{tabular*}{\textwidth}{
  @{\extracolsep{\fill}} l c *{2}{S[table-format=1.2e-2, table-number-alignment=center]} *{4}{S[table-format=2.1, table-number-alignment=center]}
}
\toprule 
\multirow{2}{*}{\textbf{Operation}} & \multirow{2}{*}{\textbf{\#}} & \multicolumn{2}{c}{\textbf{MAE [J]}} & \multicolumn{2}{c}{\textbf{MAPE [\%]}} & \multicolumn{2}{c}{\textbf{WAPE [\%]}} \\
\cmidrule{3-4}\cmidrule{5-6}\cmidrule{7-8}
& & {fwd} & {bwd} & {fwd} & {bwd} & {fwd} & {bwd} \\
\midrule
BatchNormalization & 15 & 3.02e-06 & 7.06e-06 & 7.8 & 9.2 & 8.1 & 9.4 \\
DequantizeLinear & 100 & 1.08e-07 & 2.72e-06 & 13.4 & 20.8 & 10.8 & 23.5 \\
Flatten & 87 & 1.51e-07 & 1.72e-08 & 19.6 & 5.8 & 17.9 & 5.1 \\
MaxPool & 100 & 1.05e-05 & 1.00e-05 & 62.5 & 40.7 & 40.2 & 32.0 \\
QGemm & 100 & 9.96e-07 & 9.61e-06 & 15.1 & 14.2 & 14.3 & 19.5 \\
QLinearAdd & 90 & 5.39e-05 & 1.42e-05 & 69.3 & 18.2 & 67.0 & 11.5 \\
QLinearAveragePool & 100 & 1.07e-05 & 1.09e-05 & 24.1 & 10.0 & 18.2 & 7.4 \\
QLinearConv & 100 & 6.86e-05 & 2.54e-04 & 80.1 & 60.4 & 43.3 & 11.6 \\
QuantizeLinear & 100 & 3.64e-06 & 2.86e-07 & 37.6 & 24.5 & 21.9 & 16.4 \\
Relu & 15 & 6.15e-06 & 5.06e-06 & 8.1 & 30.7 & 8.2 & 30.7 \\
Resize & 100 & 7.94e-05 & 1.71e-05 & 64.9 & 6.2 & 66.5 & 7.0 \\
\midrule
Total Layers & 907 & 2.47e-05 & 3.52e-05 & 41.8 & 22.5 & 46.9 & 11.4 \\
\textbf{Total Layers} & \textbf{1814} & \multicolumn{2}{c}{\textbf{\num{3.00e-05}}} & \multicolumn{2}{c}{\textbf{32.2}} & \multicolumn{2}{c}{\textbf{16.6}} \\
\midrule
\gls{FRAM} Read & 50 & \multicolumn{2}{c}{\num{1.51e-04}} & \multicolumn{2}{c}{0.9} & \multicolumn{2}{c}{0.9} \\
\gls{FRAM} Write & 50 & \multicolumn{2}{c}{\num{1.54e-04}} & \multicolumn{2}{c}{0.9} & \multicolumn{2}{c}{0.9} \\
\textbf{Total \gls{FRAM}} & \textbf{100} & \multicolumn{2}{c}{\textbf{\num{1.48e-04}}} & \multicolumn{2}{c}{\textbf{0.9}} & \multicolumn{2}{c}{\textbf{0.9}} \\
\bottomrule
\end{tabular*}
\end{table}

Analyzing the results further reveals a slight but consistent bias of the layer regression model underestimates the energy consumption of backward passes (which are typically more expensive due to gradient computation and checkpointing overhead), while overestimating forward passes.
This indicates minor inaccuracies resulting from the abstraction of some implementation details, e.g., handling of memory access latency or instruction pipelining, but overall confirms the linear relationship between compute, memory, and energy.

For the \gls{FRAM} regression model, we observed an almost perfect linear relationship between data size and energy consumption using 50 measurements per read and write within a range of \SI{10}{kB} to \SI{300}{kB}. 

In Table~\ref{tab:result_metrics}, we present a quantitative evaluation of the two regression models. We report three error metrics: Mean absolute error (MAE), mean absolute percentage error (MAPE), and weighted absolute percentage error (WAPE).
The WAPE was introduced to provide a more balanced assessment. Since energy consumption varies by orders of magnitude across layers (e.g., quantization versus convolutional layers), MAPE is dominated by high-energy layers.
WAPE mitigates this by weighting errors proportional to each layer's energy contribution.

To ensure a fair comparison of the layer regression model across all layer types, we picked a subset of 100 samples of each layer type for quantitative evaluation, thereby addressing the class imbalance in the original dataset.
Notably, layers such as ReLU or BatchNorm are underrepresented, as they are often fused into preceding layers and thus modeled implicitly.
The layer regression model achieves a MAPE of 32.2\,\% and a WAPE of 16.6\,\%. While this may seem like a high error at first glance, it is still sufficient for our intended use case as the model reliably distinguishes between feasible and infeasible architectures under strict intermittency constraints.
For example, it correctly identifies when a candidate architecture exceeds the per-layer energy budget thereby preventing energy underflow during training cycles.

The \gls{FRAM} regression model achieves a MAPE and WAPE of 0.9\,\%, demonstrating that persistent-state overhead can be predicted with high precision, enabling accurate planning of checkpoint intervals in intermittent execution.

\begin{figure}[t]
    \centering
    \includegraphics[width=0.7\textwidth]{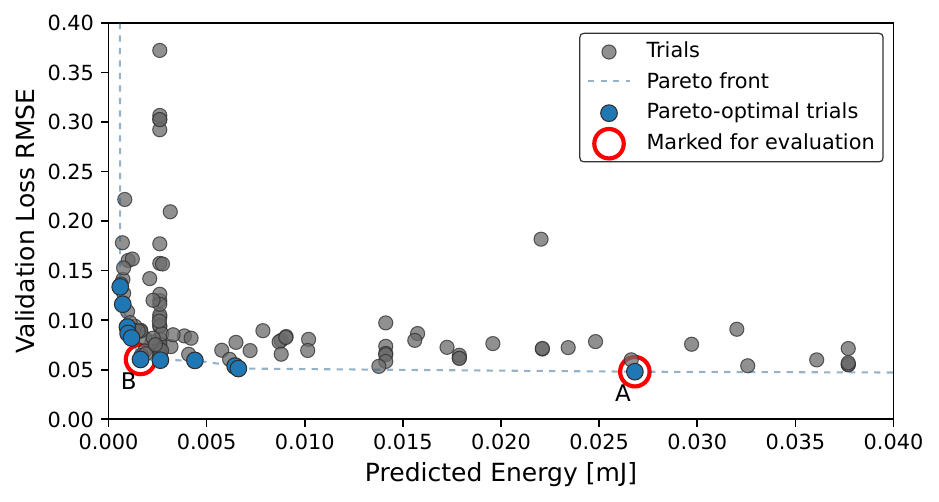}
    \caption{\gls{MOO} results for the CWRU dataset optimizing the \glspl{DNN} for energy efficiency and validation loss when utilizing the per-layer energy consumption prediction models.}
    \label{fig:results_pareto}
\end{figure}

\subsection{Optimizing Autoencoders for Intermittency Aware Execution}
Using the layer and \gls{FRAM} regression models, we demonstrate their applicability in our \gls{MOO} framework. For evaluation we selected the CWRU dataset\footnote{\url{https://engineering.case.edu/bearingdatacenter/12k-drive-end-bearing-fault-data}}, which contains bearing vibration data under normal and faulty operating conditions. We search for an optimized autoencoder architecture that can be deployed on the nrf52840 Cortex-M4 \gls{MCU} and is suitable for on-device intermittent learning. The use-case of anomaly detection is particularly well-suited for intermittent learning as autoencoders can be trained unsupervised, i.e., without any hard labels. All candidate models are convolutional autoencoders with a fixed input shape of (2\,$\times$\,256), identical encoder-decoder depth, and a latent dimension of 16. 

During search we vary the number of hidden channels and the individual pruning configuration of the convolutional layers. The \gls{MOO} minimizes both validation reconstruction loss on normal data and predicted total energy demand. We use Optuna's NSGA-II sampler for 200 trials to approximate the Pareto front. Each candidate is trained for 20 epochs using mean squared reconstruction error and the Adam optimizer. Pruning is applied at epoch 10 for all trials. 
Since the fault conditions in the CWRU dataset are clearly separable from the normal data, the validation reconstruction loss on normal data is used as the model-quality objective here, instead of anomaly detection metrics like AUROC.

The Pareto front resulting from the optimization can be seen in Fig.~\ref{fig:results_pareto}. It shows the trade-off between validation reconstruction loss and predicted total energy demand. Comparing point A, which achieves the lowest validation loss, with point B, which accepts a slightly higher loss, the predicted total energy demand decreases from \SI{0.0268}{mJ} to \SI{0.00163}{mJ}. The predicted energy demand of the most expensive layer decreases from \SI{0.0198}{mJ} to \SI{0.000449}{mJ}. Despite this substantial reduction in predicted energy demand, both points correctly separate all anomalous test samples from normal operating conditions.
\section{Conclusion}

We presented an approach to enable on-device training of \gls{DNN} in energy-harvesting environments using \gls{IC}. To this end, we proposed a prediction model that can estimate per-layer energy consumption for both inference and training. We combined the prediction model with a hardware-aware \gls{MOO} algorithm, thereby allowing for intermittency aware search of efficient \gls{DNN} architectures for deployment on energy harvesting \gls{MCU}.
We demonstrated the capabilities of the approach using an anomaly detection use case deployed on a nRF52840 Cortex-M4 \gls{MCU}, where our prediction model could reach 16.6\,\% of WAPE in energy prediction for network layers and 0.9\,\% for \gls{FRAM} state preservation in the design stage.
As a result, we were able to achieve an energy decrease of around 94\,\% with only minimal increase in validation loss by selecting a more efficient \gls{DNN} with our approach.

\begin{credits}
\subsubsection{\ackname}
This work was partially funded by the Deutsche Forschungsgemeinschaft 
(DFG, German Research Foundation) -- project number 539710462 (``DOSS'') 
and from the Bundesministerium für Forschung, Technologie und Raumfahrt 
(BMFTR, Federal Ministry of Research, Technology and Space) in Germany 
for the project SUSTAINET-inNOvAte 16KIS2262.

\subsubsection{\discintname}
The authors have no competing interests to declare that are
relevant to the content of this article.
\end{credits}

\bibliographystyle{splncs04}
\bibliography{references}

\end{document}